\documentclass[a4paper]{styles/svproc}

\usepackage{url}

\usepackage{amsmath,amssymb,amsfonts}
\usepackage{graphicx}
\usepackage{placeins}
\usepackage{textcomp}
\usepackage{xcolor}
\usepackage{acronym}
\usepackage{float}

\acrodef{MPC}{Model Predictive Control}
\acrodef{QP}{quadratic program}
\acrodef{DOF}{degree-of-freedom}
\acrodef{RMS}{root mean square}

\begin{document}
\mainmatter

\title{Constraint-Unified MPC for Over-Actuated Surface Vehicles with Post-Detection Fault Reconfiguration}
\titlerunning{Constraint-Unified MPC for Over-Actuated Surface Vehicles}

\author{
        Sebastian Burmester 
 \and   Ruiheng Jiang 
 \and   Noa Sendlhofer \and \\
        Raffaello D'Andrea
 \and   Aswin Ramachandran
 }
\authorrunning{S. Burmester et al.}
\tocauthor{Sebastian Burmester, Ruiheng Jiang, Noa Sendlhofer, Raffaello D'Andrea, Aswin Ramachandran}

\institute{Institute for Dynamic Systems and Control (IDSC),\\
ETH Z\"urich, Switzerland,\\
\email{sburmester@ethz.ch, aramachandra@ethz.ch}}

\maketitle

\begin{abstract}
Choreographed aquatic performances require small
autonomous surface vehicles to track precise paths under per-thruster force and rate limits, including after thruster failures.
We report a deployed system in which trajectory tracking, thrust allocation, and the per-thruster force and rate limits are resolved in a single quadratic program over the per-thruster commands, with fault reconfiguration entering through one binary flag per thruster from an external detector.
The system has driven a fleet in live performances on Lake Z\"urich and at the Time Space Existence 2025 exhibition in Venice.
The field campaign measures $1.6\,\mathrm{cm}$ root mean square position error in a $10$-minute hold and $4.3\,\mathrm{cm}$ over a $10\,\mathrm{m}$ square at $0.6\,\mathrm{m/s}$. 
At $0.5\,\mathrm{m/s}$, losing the front thruster increases the error to $11.6\,\mathrm{cm}$, while losing the starboard-side thruster increases it to $11.5\,\mathrm{cm}$.
Losing two thrusters simultaneously leaves the craft tracking a $0.4\,\mathrm{m/s}$ square with a root mean square position error of $1.25\,\mathrm{m}$.
Compared to our own cascaded baseline, the unified formulation tracks the nominal square to the same few centimeters and holds station more tightly with roughly half the thruster force. The architectures separate after a thruster failure, where the unified controller stays within $24\,\mathrm{cm}$ of the reference path while the cascade leaves it.
\end{abstract}

\begin{figure}[h]
    \centering
    \includegraphics[width=0.83\linewidth]{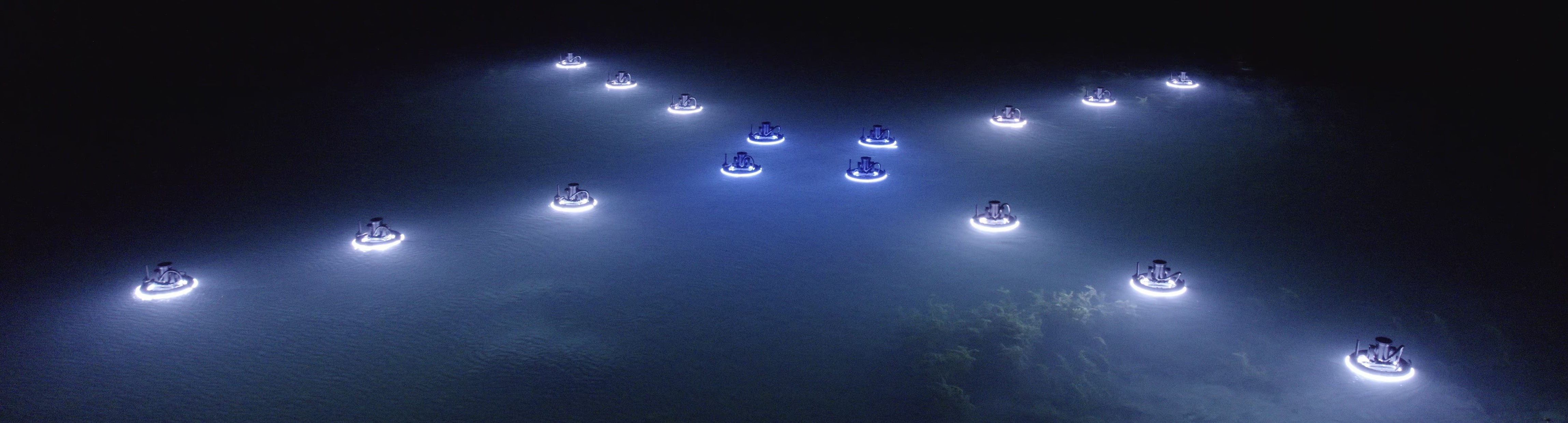}
    \caption{Live performance on Lake Z\"urich}
    \label{fig:intro_swarm}
\end{figure}

\section{Introduction}

At dusk, a swarm of autonomous surface vehicles glides into formation on a lake, ready for a live choreography. The craft must track choreographed paths to centimeter accuracy~\cite{wayofwater_nime2026,kamm2026motion,jiang2026crossplatform} under per-thruster force and rate limits, and must keep tracking them when a thruster fails mid-performance.

Each craft is an over-actuated four-thruster omnidirectional vehicle whose onboard \ac{MPC} loop solves one condensed \ac{QP} over the per-thruster commands at $10\,\mathrm{Hz}$. Tracking, thrust allocation, per-thruster force and rate limits, and post-detection outage accommodation all appear in that single program. The system has driven a fleet on Lake Z\"urich and at the Time Space Existence 2025 exhibition in Venice~\cite{wayofwater_nime2026} (Fig.~\ref{fig:intro_swarm}). A conventional cascaded architecture instead separates these stages, the tracker requesting a body-frame generalized force before allocation and thrust limits. Both architectures receive the same fault flag and rebuild their allocation over the surviving thrusters. Fault detection itself remains external to both.

Our contributions are \emph{(i)} a unified condensed \ac{QP} over the per-thruster
commands, folding trajectory tracking, thrust allocation, per-thruster force and rate limits, and post-detection fault reconfiguration into one program, \emph{(ii)} its deployment in live performances and evaluation on one craft under repeatable field conditions, and \emph{(iii)} a paired field comparison against a conventional LQR cascade developed with comparable design and tuning effort.

\section{Related Work}\label{sec:related_work}

\subsection{Conventional Cascaded Architecture} \label{sec:conventional_pipeline}
Marine heading controllers include PID~\cite{wang2024pid_usv} and fuzzy-LQR~\cite{yazdanpanah2013fuzzy_lqr}. A station-keeping study compares PD, backstepping, and sliding-mode controllers that compute generalized forces before thrust allocation~\cite{sarda2016stationkeeping}.

We compare against a baseline in this style: a tracker, a pseudoinverse allocator, and a downstream limiter, with the allocator rebuilt over surviving channels after a declared failure~\cite{cristofaro2014ftc_alloc}. Actuator feasibility therefore sits downstream of tracking, unlike constrained allocation methods that also enforce bounds and rate limits inside the allocation problem~\cite{harkegard2003allocation,johansen2013allocation}.

\subsection{Fault Handling and Robust MPC} \label{sec:related_aftc}
Marine \ac{MPC} studies include work on under-actuated vessels~\cite{oh2010los_mpc_underact,yan2012mpc_rnn_underact,abdelaal2018nmpc_underact,rossides2022mhenmpc}, which differ from the over-actuated platform considered here. Nonlinear \ac{MPC} has been demonstrated in real-world experiments~\cite{wang2020roboat2} and in under-actuated tracking simulations~\cite{liu2015nmpc_usv}.

Active fault-tolerant control generally combines diagnosis with reconfiguration~\cite{zhang2003ftc}. Marine examples include residual-based diagnosis with redundant-actuator accommodation~\cite{corradini2011ftc_rov} and observer-based isolation with control reallocation~\cite{cristofaro2014ftc_alloc}.

For heading control, an adaptive Kalman filter estimates actuator-fault parameters~\cite{tandfonline2025aftc}. Nonlinear \ac{MPC} combined with moving horizon estimation of two azimuth-thruster efficiencies supports tracking and fault detection~\cite{rossides2022mhenmpc}. Another approach isolates faults by reweighting the \ac{MPC} objective toward a diagnostic azimuth angle while retaining vessel dynamics and input constraints in the optimization~\cite{tsolakis2024activefdi}. Here, an external mechanism has already declared a complete thruster outage on an over-actuated platform. Detection sits outside the controller by design: many modern electronic speed controllers report per-thruster telemetry, so no dedicated sensing is needed to declare an outage~\cite{hanai2009thruster_fdd}.

\section{System Model and Problem Formulation}
\label{sec:problem_formulation}
Live aquatic shows require autonomous surface vehicles to follow choreographed paths precisely, while accommodating actuator failures gracefully. We first introduce the vehicle model and then specify the problem.

\subsection{Vehicle Model}
\label{sec:vehicle_model}
\begin{figure}[!htbp]
    \centering
    \includegraphics[width=\linewidth]{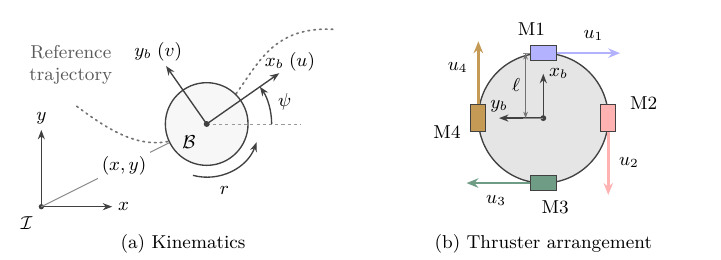}
    \caption{Vehicle coordinates and actuation.}
    \label{fig:vehicle_kinematics}
\end{figure}

The controller was validated on a small omnidirectional autonomous surface vehicle with four bidirectional thrusters placed symmetrically about the center of gravity at equal moment arms (Fig.~\ref{fig:vehicle_kinematics}(b)). RTK-GPS position and IMU angular-rate measurements provide the observations used to estimate the vehicle state.

We adopt the 3-degree-of-freedom horizontal-plane Fossen model~\cite{fossen2011marine}. We assume that the body-fixed frame $\mathcal{B}$ is centered on the vehicle and aligned with the center of mass, the center of buoyancy, and the center of flotation. Figure~\ref{fig:vehicle_kinematics}(a) defines the inertial and body frames. The vessel pose is $\boldsymbol{\eta} = [x,\ y,\ \psi]^T$ in the inertial frame, and its body-fixed velocities are $\boldsymbol{\nu} = [u,\ v,\ r]^T$. Together they form the state $\mathbf{x}_k = [\boldsymbol{\eta}_k^T, \boldsymbol{\nu}_k^T]^T \in \mathbb{R}^6$. The kinematics satisfy $\dot{\boldsymbol{\eta}} = \mathbf{J}(\psi)\boldsymbol{\nu}$ with
\begin{equation}
\mathbf{J}(\psi) = \begin{bmatrix} \cos\psi & -\sin\psi & 0 \\ \sin\psi & \cos\psi & 0 \\ 0 & 0 & 1 \end{bmatrix},
\end{equation}
and the dynamics
\begin{equation}
\mathbf{M}\dot{\boldsymbol{\nu}} + \mathbf{N}(\boldsymbol{\nu})\boldsymbol{\nu} = \boldsymbol{\tau}_c,
\end{equation}
where $\boldsymbol{\tau}_c = [X,\ Y,\ N]^T$ is the body-frame generalized force. Combining rigid-body and added-mass inertia gives a diagonal inertia matrix $\mathbf{M} = \mathrm{diag}\{m - X_{\dot{u}},\ m - Y_{\dot{v}},\ I_z - N_{\dot{r}}\}$, and combining rigid-body Coriolis-centripetal terms and linear hydrodynamic damping yields
\begin{equation}
\label{eq:vehicle_coupling}
\mathbf{N}(\boldsymbol{\nu}) = \begin{pmatrix}
-X_u & -mr & Y_{\dot{v}}v \\
mr & -Y_v & -X_{\dot{u}}u \\
-Y_{\dot{v}}v & X_{\dot{u}}u & -N_r
\end{pmatrix}.
\end{equation}

The thruster arrangement in Fig.~\ref{fig:vehicle_kinematics}(b) determines the allocation matrix $\mathbf{T}$. The commands $\mathbf{u}_k = [u_{1,k}, u_{2,k}, u_{3,k}, u_{4,k}]^T \in \mathbb{R}^4$ produce the body-frame generalized forces at sampling step $k$. With $\ell$ the thruster moment arm about the vehicle center,
\begin{equation}
\label{eq:allocation}
\boldsymbol{\tau}_{c,k} = \mathbf{T}\, \mathbf{u}_k = \begin{bmatrix} 0 & -1 & 0 & 1 \\ -1 & 0 & 1 & 0 \\ -\ell & -\ell & -\ell & -\ell \end{bmatrix} \begin{bmatrix} u_{1,k} \\ u_{2,k} \\ u_{3,k} \\ u_{4,k} \end{bmatrix} = \begin{bmatrix} X_k \\ Y_k \\ N_k \end{bmatrix}.
\end{equation}

Each bidirectional thruster has a command bound $-F_{\max} \le u_{i,k} \le F_{\max}$. Successive commands also satisfy $|u_{i,k} - u_{i,k-1}| \le \Delta u_{\max}$, where $\Delta u_{\max}$ is the maximal change per time step. Table~\ref{tab:parameters} lists the vehicle parameters and command limits.

\begin{table}[!htbp]
    \centering
    \footnotesize
    \setlength{\tabcolsep}{4pt}
    \caption{Vehicle and actuator parameters.}
    \label{tab:parameters}
    \begin{tabular}{llc}
    \hline
    Mass                        & $m$                              & $30.0\,\mathrm{kg}$ \\
    Added mass                  & $X_{\dot{u}} = Y_{\dot{v}}$      & $-12.0\,\mathrm{kg}$ \\
    Linear damping              & $X_u = Y_v$                      & $-65.0\,\mathrm{N\,s\,m^{-1}}$ \\
    Yaw damping                 & $N_r$                            & $-3.5$ / $-0.35\,\mathrm{N\,m\,s\,rad^{-1}}$ \\
    Combined yaw inertia        & $J_{\text{comb}}$                & $2.0$ / $0.20\,\mathrm{kg\,m^2}$ \\
    Thruster moment arm         & $\ell$                           & $0.25\,\mathrm{m}$ \\
    \hline
    Thruster command bound      & $F_{\max}$                       & $25\,\mathrm{N}$ \\
    Rate bound                  & $\Delta u_{\max}$                & $3\,\mathrm{N}$ per step \\
    \hline
    \end{tabular}
\end{table}
\FloatBarrier

\subsection{Tracking Task}
\label{sec:actuation_task}

Given a reference state trajectory $\mathbf{x}_{\text{ref},k}$ and the estimated current state, the task is to track the reference while respecting the available actuation.

Thruster loss changes the available forces and moments, so nominal tracking commands may no longer be achievable. The task therefore requires robust fault reconfiguration that redistributes commands among the surviving thrusters while respecting their force and rate limits. An external detector supplies one binary flag per thruster. The set $\mathcal{F}\subseteq\{1,2,3,4\}$ contains the declared complete outages, for which no thrust is available. Declarations may arrive after the physical outage, as in the delayed-flag experiments of Sect.~\ref{sec:results}.

\section{Methodology}
\label{sec:methodology}
We first present the consolidated \ac{MPC} problem, including its decision variables, cost, and fault reconfiguration. We then describe the prediction model, condensed \ac{QP} and real-time implementation, and the cascaded baseline used in the experiments.

\subsection{Consolidated MPC Problem}
\label{sec:consolidated_mpc}
\begin{figure}[!htbp]
    \centering
    \includegraphics[width=\linewidth]{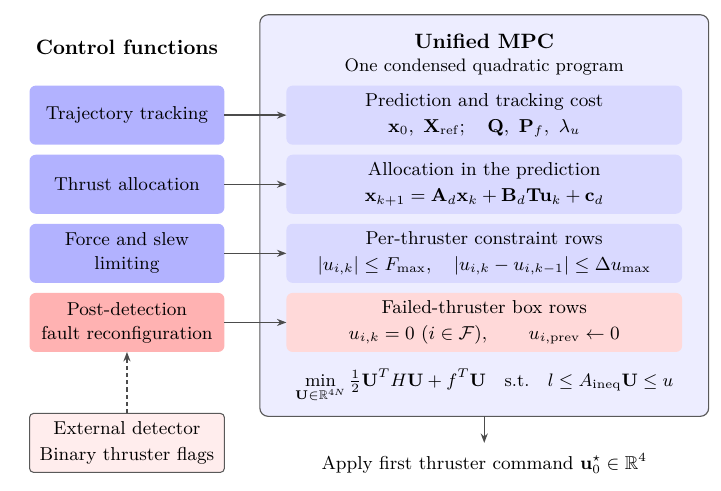}
    \caption{Mapping four control functions into one condensed QP.}
    \label{fig:unified_architecture}
\end{figure}

We combine the four functions of Sect.~\ref{sec:conventional_pipeline} by optimizing the per-thruster commands $\mathbf{u}_k \in \mathbb{R}^4$ directly (Fig.~\ref{fig:unified_architecture}). Allocation enters prediction through $\mathbf{B}_d\mathbf{T}$, tracking and input effort enter the cost, and force, slew, and declared-outage limits enter the constraints.

Linearizing the vehicle model and discretizing with period $T_s$ gives the predictor $(\mathbf{A}_d, \mathbf{B}_d, \mathbf{c}_d)$ described in Sect.~\ref{sec:prediction_model}. Over a horizon of $N$ steps, state and input deviations are $\tilde{\mathbf{x}}_k = \mathbf{x}_k - \mathbf{x}_{\text{ref},k}$ and $\tilde{\mathbf{u}}_k = \mathbf{u}_k - \mathbf{u}_{\text{ref},k}$, with zero input reference $\mathbf{u}_{\text{ref},k}=\mathbf{0}$, so the input term reduces to a pure effort regularization $\lambda_u|\mathbf{u}_k|^2$ rather than tracking of a feedforward command. When a thruster is declared failed, its entry in the previously applied input $\mathbf{u}_{\text{prev}}$ is reset to zero before constraint assembly. At each sampling instant, we solve
\begin{equation}
\label{eq:mpc_problem}
\begin{aligned}
\min_{\mathbf{u}} \quad & \sum_{k=0}^{N-1} \left( \tilde{\mathbf{x}}_k^T \mathbf{Q}\, \tilde{\mathbf{x}}_k + \tilde{\mathbf{u}}_k^T \mathbf{R}\, \tilde{\mathbf{u}}_k \right) + \tilde{\mathbf{x}}_N^T \mathbf{P}_f\, \tilde{\mathbf{x}}_N \\
\text{s.t.}\quad & \mathbf{x}_{k+1} = \mathbf{A}_d \mathbf{x}_k + \mathbf{B}_d \mathbf{T} \mathbf{u}_k + \mathbf{c}_d, && k = 0,\dots,N-1, \\
                 & -F_{\max} \le u_{i,k} \le F_{\max}, && \forall i,\ k, \\
                 & |u_{i,k} - u_{i,k-1}| \le \Delta u_{\max}, && \forall i,\ k \ge 1, \\
                 & |u_{i,0} - u_{i,\text{prev}}| \le \Delta u_{\max}, && \forall i, \\
                 & u_{i,k} = 0, && \forall k,\ i \in \mathcal{F}, \\
                 & \mathbf{x}_0 = \mathbf{x}(t_k).
\end{aligned}
\end{equation}

The per-step commands are stacked into the decision variable
\begin{equation}
    \mathbf{u} = [\mathbf{u}_0^T, \mathbf{u}_1^T, \ldots, \mathbf{u}_{N-1}^T]^T \in \mathbb{R}^{4N},
\end{equation}
where $u_{i,k}$ commands thruster $i$ at step $k$. The dynamics in \eqref{eq:mpc_problem} map this sequence to the predicted states, using the allocation matrix of \eqref{eq:allocation}.

The cost determines how the available thrust is used for tracking. It weights state and input deviations by $\mathbf{Q}$ and $\mathbf{R}$, with $\mathbf{P}_f$ penalizing terminal state error. We use
\begin{align}
\mathbf{Q} &= \mathrm{diag}(q_{xy}, q_{xy}, q_\psi, q_{uv}, q_{uv}, q_r), \\
\mathbf{R} &= \lambda_u\, \mathbf{I}_4,
\end{align}
where $q_{xy}$ equally weights position errors, $q_{uv}$ equally weights body-fixed translational velocities, and $\lambda_u$ applies equally to all four thrusters. The state weights set the tracking priorities under degraded actuation: raising $q_{xy}$ relative to $q_\psi$ prioritizes position over heading. We use $\mathbf{P}_f = \mathbf{Q}$, so no formal stability guarantee is established and stability is evaluated empirically in field experiments.

The previous-input reset keeps the failed-thruster zero-command constraint compatible with the first-step rate limit in \eqref{eq:mpc_problem}.

\subsection{Prediction Model}
\label{sec:prediction_model}
The nonlinear dynamics $\dot{\mathbf{x}} = f(\mathbf{x}, \boldsymbol{\tau}_c)$ are linearized about $(\mathbf{x}_0, \boldsymbol{\tau}_{c,0})$. Here, $\mathbf{x}_0$ is the latest measured state and $\boldsymbol{\tau}_{c,0} = \mathbf{T}\mathbf{u}_{\text{prev}}$ is the generalized force of the previously applied input. This gives $\dot{\mathbf{x}} \approx \mathbf{A}_c \mathbf{x} + \mathbf{B}_c \boldsymbol{\tau}_c + \mathbf{c}_c$, with $\mathbf{A}_c$ and $\mathbf{B}_c$ the Jacobians of $f$ at that point and $\mathbf{c}_c = f_0 - \mathbf{A}_c \mathbf{x}_0 - \mathbf{B}_c \boldsymbol{\tau}_{c,0}$ the affine offset, which is non-zero because the operating point is not necessarily an equilibrium.

The model is discretized exactly over $T_s$ by a single exponential of the augmented generator built from $(\mathbf{A}_c, \mathbf{B}_c, \mathbf{c}_c)$ acting on $\xi = [\mathbf{x}^T, \boldsymbol{\tau}_c^T, 1]^T$. Its upper block row gives the zero-order-hold triple $(\mathbf{A}_d, \mathbf{B}_d, \mathbf{c}_d)$ directly, and the discrete offset $\mathbf{c}_d$ is folded into the \ac{QP} gradient below.

Substituting $\boldsymbol{\tau}_{c,k} = \mathbf{T}\mathbf{u}_k$ gives the predictor in \eqref{eq:mpc_problem} in four thruster commands, so constraints act directly on the decision variable. The model remains fixed over the horizon and is re-linearized once per control step.

\subsubsection{Two Yaw Parameter Sets.}
The yaw parameter pairs in Table~\ref{tab:parameters} give the system-identified plant and estimator values first and deployed \ac{MPC} predictor values second. Combined yaw inertia is $J_{\text{comb}} = I_z - N_{\dot r}$, the third diagonal entry of $\mathbf{M}$ in Sect.~\ref{sec:vehicle_model}. The plant assumes instantaneous delivery of commanded generalized force, while physical thrusters have actuator dynamics. The predictor reduces the magnitudes of yaw damping $N_r$ and combined yaw inertia $J_{\text{comb}}$ by a factor of ten, preserving the yaw pole,
$N_r/J_{\text{comb}} = -1.75\,\mathrm{s^{-1}}$, while increasing the input gain $1/J_{\text{comb}}$ tenfold. This simplified, platform-specific adjustment is not an identified thruster-delay model, and physical lag is not estimated here. A later controller uses a delay-compensated predictor, outside this paper's experiments and evaluation.

\subsection{Condensed QP and Real-Time Implementation}
\label{sec:qp_implementation}
Successive substitution eliminates the states from the discrete dynamics~\cite[Sect.~8.8.4]{rawlings2017model}, giving
\begin{equation}
\mathbf{X} = \mathcal{S}_x \mathbf{x}_0 + \mathcal{S}_u \mathbf{U} + \mathcal{S}_c,
\end{equation}
where $\mathbf{X} = [\mathbf{x}_1^T, \ldots, \mathbf{x}_N^T]^T$ and $\mathbf{U}$ denotes the stacked decision variable $\mathbf{u}\in\mathbb{R}^{4N}$ of Sect.~\ref{sec:consolidated_mpc}. The matrices $\mathcal{S}_x$ and $\mathcal{S}_u$ propagate the initial state and inputs, respectively, with $\mathbf{B}_d\mathbf{T}$ incorporating allocation and $\mathcal{S}_c$ accumulating affine drift.

Substituting this prediction into the cost yields
\begin{equation}
\min_{\mathbf{U}} \tfrac{1}{2} \mathbf{U}^T H\, \mathbf{U} + f^T \mathbf{U} \quad \text{s.t.}\quad l \le A_{\text{ineq}} \mathbf{U} \le u,
\end{equation}
with $H = 2 ( \mathcal{S}_u^T \bar{\mathbf{Q}}\, \mathcal{S}_u + \lambda_u\, \mathbf{I}_{4N} )$ and $f = 2\, \mathcal{S}_u^T \bar{\mathbf{Q}}\, (\mathcal{S}_x \mathbf{x}_0 + \mathcal{S}_c - \mathbf{X}_{\text{ref}})$, where $\bar{\mathbf{Q}} = \mathrm{blkdiag}(\mathbf{Q}, \ldots, \mathbf{Q}) \in \mathbb{R}^{6N \times 6N}$ stacks the stage cost matrix. They are assembled from the current prediction model and weights, with $f$ accounting for the initial state, affine drift, and reference.

The matrix $A_{\text{ineq}}$ stacks box, successive-input-difference, and first-step rows. The first-step bounds are $[u_{i,\text{prev}} - \Delta u_{\max},\, u_{i,\text{prev}} + \Delta u_{\max}]$, updated from the previously applied input. On a fault declaration, the failed thruster's box bounds are tightened to zero and its previous-input entry is simultaneously reset to zero. This prevents conflict between its zero-command constraint and the rate window inherited from a loaded actuator.

State estimation uses an extended Kalman filter that fuses RTK-GPS position ($7\,\mathrm{Hz}$) and IMU angular rate ($100\,\mathrm{Hz}$). At $f_s = 10\,\mathrm{Hz}$, the controller reads the latest estimated state, builds the \ac{QP}, and calls OSQP~\cite{stellato2020osqp}.

We use a prediction horizon $N=10$, stage cost $\mathbf{Q}=\mathrm{diag}(4,4,6,1,1,6)$, terminal cost $\mathbf{P}_f=\mathbf{Q}$, and input weight $\lambda_u=10^{-4}$. OSQP uses absolute and relative tolerances $\varepsilon_{\text{abs}} = \varepsilon_{\text{rel}} = 10^{-4}$. The weights $\mathbf{Q}$ and $\lambda_u$ were selected by simulation-based grid search and validated in the field (Sect.~\ref{sec:results}).

\subsection{Cascaded Baseline}
\label{sec:baseline}
The cascade of Sect.~\ref{sec:conventional_pipeline} uses the same plant, reference, actuator limits, and metrics as the \ac{MPC}. Its tracker is an infinite-horizon LQR re-solved online: at each step it reuses the zero-order-hold model $(\mathbf{A}_d, \mathbf{B}_d, \mathbf{c}_d)$ formed about the same operating point $(\mathbf{x}_0, \mathbf{T}\mathbf{u}_{\text{prev}})$ that the \ac{MPC} predicts from, and solves the discrete-time algebraic Riccati equation for that model. The tracker is therefore the LQ counterpart of the \ac{MPC}'s own successive-linearization predictor, which is intended to leave the placement of the actuator limits as the main difference between the two architectures. We match the weights as far as the two formulations allow: $\mathbf{Q}$ is identical, and the input weight is transformed as $\mathbf{R} = \lambda_u \mathbf{G}^{T}\mathbf{G}$, with $\mathbf{G}$ the active allocation matrix, so that the penalty on generalized force corresponds as closely as possible to the \ac{MPC}'s penalty on the individual thruster commands and control effort is not underpriced in particular directions after a failure. Tracking uses standard finite preview~\cite{tomizuka1975preview} over the same $N$ reference steps as the \ac{MPC} horizon.

Failure is declared by the same external flag. On declaration, the allocator is rebuilt over the healthy thruster set $\mathcal{H}$ as $\mathbf{G} = (\mathbf{T}_{\mathcal{H}})^{+}$, with failed outputs held at zero~\cite{cristofaro2014ftc_alloc}. The input weight $\mathbf{R}$ updates alongside $\mathbf{G}$ to price the actually available thrust. Saturation is applied downstream in sequence: first the rate limit $\Delta u_{\max}$ relative to $\mathbf{u}_{\text{prev}}$, then the absolute box limit $F_{\max}$. Reconfiguration is thus confined to the allocation layer~\cite{harkegard2003allocation}. The tracker knows nothing of actuator limits and the fault reaches it only through the resulting tracking error.

\begin{samepage}
\section{Experimental Validation} \label{sec:results}

The experiments are organized into four sets: station-hold and nominal-square runs, immediate-detection square runs with one or two thrusters disabled at motion onset, delayed mid-lap failure runs, and paired runs with the cascaded baseline.

\end{samepage}

\subsection{Experimental Setup and Metrics}
\label{sec:experimental_setup}

Experiments were conducted on Lake Z\"urich in around $7\,\mathrm{km/h}$ wind, with $2$--$5\,\mathrm{cm}$ short-crested ripples. Two references are used: a ten-minute static pose hold and a $10\,\mathrm{m}$ rounded square ($\approx 2\,\mathrm{m}$ corner radius) with heading held tangent to the path. The commanded speed is set per run: the nominal square uses $0.6\,\mathrm{m/s}$, the single-thruster runs use $0.5\,\mathrm{m/s}$, and the simultaneous-loss runs use $0.4\,\mathrm{m/s}$.

For immediate-detection square runs, a scheduler zeros the selected thruster command while the autonomous surface vehicle is stationary, with the external flag available before motion. Delayed mid-lap runs apply the same command-level outage but withhold the flag for approximately two seconds. Table~\ref{tab:error_metrics} covers station-hold, nominal-square, and immediate-detection runs, including front- (\textit{F}), starboard-side- (\textit{S}), and simultaneous front-plus-starboard-side-failure (\textit{F+S}) cases. Paired cascade rows use the same conditions.

For the rows in Table~\ref{tab:error_metrics}, the position error is the horizontal Euclidean error magnitude, and the heading error is wrapped before computing its \ac{RMS}. The \ac{RMS}, $p_{95}$, maximum, and saturation fraction use ten minutes for the station hold and one lap for each square run. The saturation fraction counts thruster samples within $0.1\%$ of the software command bounds, $\pm F_{\max}$.

\begin{table}[H]
    \centering
    \footnotesize
    \setlength{\tabcolsep}{5pt}
    \caption{Position and heading errors for the station-hold, nominal-square, and immediate-detection square experiments.}
    \label{tab:error_metrics}
    \newsavebox{\errormetricstable}
    \begin{lrbox}{\errormetricstable}
    \begin{tabular}{llrrrrr}
    \hline
    Run & Controller & \ac{RMS} & $p_{95}$ & Max & $\psi$ \ac{RMS} & Sat. \\
        &            & [m]      & [m]      & [m] & [deg]           & [\%] \\
    \hline
    Hold, $10\,$min          & \ac{MPC} & $0.016$ & $0.027$ & $0.074$ & $4.2$   & $0.0$  \\
    Hold, $10\,$min          & cascade  & $0.030$ & $0.057$ & $0.131$ & $4.0$   & $0.0$  \\
    Nom., $0.6\,$m/s         & \ac{MPC} & $0.043$ & $0.076$ & $0.108$ & $2.9$   & $0.8$  \\
    Nom., $0.6\,$m/s         & cascade  & $0.047$ & $0.092$ & $0.123$ & $2.7$   & $2.3$  \\
    F, $0.5\,$m/s            & \ac{MPC} & $0.116$ & $0.194$ & $0.229$ & $24.3$  & $0.2$  \\
    F, $0.5\,$m/s            & cascade  & $5.937$ & $9.376$ & $9.830$ & $101.7$ & $31.0$ \\
    S, $0.5\,$m/s            & \ac{MPC} & $0.115$ & $0.160$ & $0.233$ & $55.6$  & $0.5$  \\
    S, $0.5\,$m/s            & cascade  & $6.090$ & $9.666$ & $10.290$ & $106.8$ & $29.0$ \\
    F+S, $0.4\,$m/s          & \ac{MPC} & $2.726$ & $5.288$ & $5.405$ & $57.6$  & $5.8$  \\
    F+S, $0.4\,$m/s, $q_\psi = q_r = 0$ & \ac{MPC} & $1.249$ & $2.028$ & $2.210$ & $153.3$ & $18.4$ \\
    \hline
    \end{tabular}
    \end{lrbox}
    \usebox{\errormetricstable}
    \par\smallskip
    \begin{minipage}{\wd\errormetricstable}
    \footnotesize\raggedright
    \end{minipage}
\end{table}

\subsection{Tracking Results}
\label{sec:tracking_results}

\subsubsection{Station Keeping and Nominal Tracking.}
\begin{figure}[!ht]
    \centering
    \includegraphics[width=\linewidth]{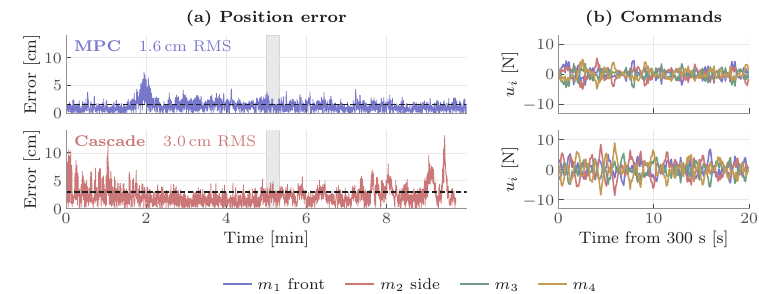}
    \caption{Station keeping position error and per-thruster commands.}
    \label{fig:hold_error}
    \par\smallskip
\end{figure}

Over the ten-minute station hold, the autonomous surface vehicle has a position-error \ac{RMS} of $1.6\,\mathrm{cm}$, a $95$th-percentile error of $2.7\,\mathrm{cm}$, and a maximum error of $7.4\,\mathrm{cm}$ (Fig.~\ref{fig:hold_error}). On the nominal square at $0.6\,\mathrm{m/s}$, the position-error \ac{RMS} is $4.3\,\mathrm{cm}$ and the maximum error is $10.8\,\mathrm{cm}$ (Table~\ref{tab:error_metrics}, Figs.~\ref{fig:tracking_xy} and~\ref{fig:tracking_actuation}).

\subsubsection{Single-Thruster Loss.}
With immediate detection before motion, losing the front thruster gives a position-error \ac{RMS} of $11.6\,\mathrm{cm}$ and starboard-side loss gives $11.5\,\mathrm{cm}$ (Table~\ref{tab:error_metrics}) on a square at $0.5\,\mathrm{m/s}$. Trajectories remain close to the reference (Fig.~\ref{fig:tracking_xy}), with commands in Fig.~\ref{fig:tracking_actuation}. The position-error \ac{RMS} is $2.7$ times the nominal-square value, while maximum errors remain below $23.3\,\mathrm{cm}$. Saturation fractions are $0.2\%$ and $0.5\%$, respectively. Heading-error \ac{RMS} increases from $2.9^\circ$ in the nominal square to $24^\circ$ for front loss and $56^\circ$ for side loss. With the nominal stage cost and reduced actuation, the controller preserves position while accepting increased heading error in these runs.

\subsubsection{Simultaneous Loss of Two Thrusters.}
\begin{figure}[!ht]
    \centering
    \includegraphics[width=\linewidth]{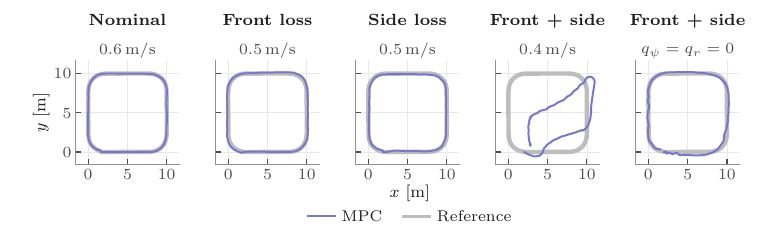}
    \caption{Measured trajectories for nominal and immediate-detection square experiments.}
    \label{fig:tracking_xy}
    \par\smallskip
\end{figure}

\begin{figure}[!ht]
    \centering
    \includegraphics[width=\linewidth]{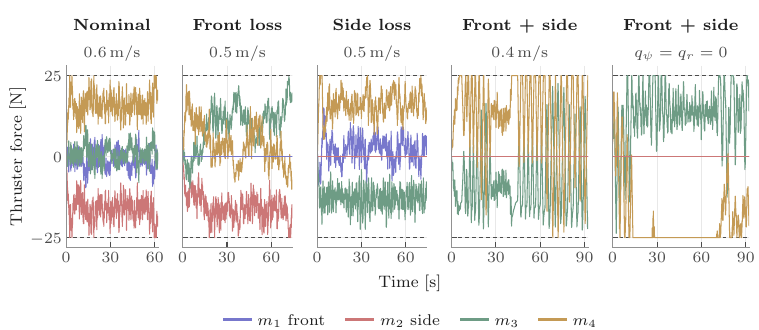}
    \caption{Per-thruster commands for nominal and immediate-detection square runs.}
    \label{fig:tracking_actuation}
    \par\smallskip
\end{figure}

Disabling the front and starboard-side thrusters together at $0.4\,\mathrm{m/s}$ gives a position-error \ac{RMS} of $2.73\,\mathrm{m}$ with the nominal weights (Table~\ref{tab:error_metrics}). The trajectory departs substantially from the square reference (Figs.~\ref{fig:tracking_xy} and~\ref{fig:tracking_actuation}).

Releasing the orientation weights ($q_\psi = q_r = 0$) reduces the position-error \ac{RMS} to $1.25\,\mathrm{m}$ and the maximum error from $5.40$ to $2.21\,\mathrm{m}$. The recorded trajectory retains the square-like shape, with increased use of surviving thrusters (Figs.~\ref{fig:tracking_xy} and~\ref{fig:tracking_actuation}). Its $153.3^\circ$ heading-error \ac{RMS} quantifies the trade-off accepted to reduce position error. After $75\,\mathrm{s}$, the released-weight run settles to $0.52\,\mathrm{m}$ along-track error and $0.09\,\mathrm{m}$ cross-track error. Over the same window, the nominal-weight run has $2.39\,\mathrm{m}$ cross-track error out of a $2.49\,\mathrm{m}$ position-error magnitude.

\subsubsection{Delayed Mid-Lap Failure under Load.}
\label{sec:fault_onset}
Four runs inject a command-level outage mid-lap at $0.5\,\mathrm{m/s}$, on a straight and a corner for each of the front and starboard-side thrusters. The external flag is withheld for approximately two seconds to represent detector latency.

\begin{figure}[!ht]
    \centering
    \includegraphics[width=\linewidth]{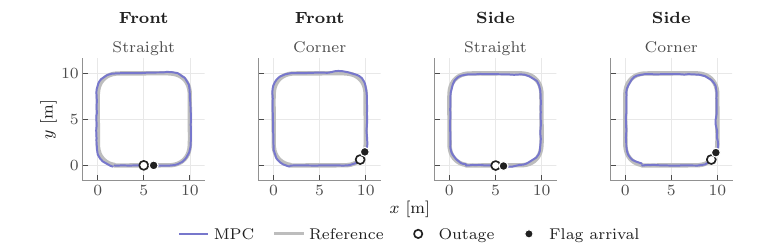}
    \caption{Measured trajectories for four delayed mid-lap outages.}
    \label{fig:fault_onset}
    \par\smallskip
\end{figure}

All four trajectories remain close to the reference (Fig.~\ref{fig:fault_onset}). During the first $15\,\mathrm{s}$ after outage injection, maximum position-error excursions are $13.4\,\mathrm{cm}$ and $20.6\,\mathrm{cm}$ for the front-thruster straight and corner cases, and $22.4\,\mathrm{cm}$ and $29.4\,\mathrm{cm}$ for the corresponding starboard-side cases. $20\,\mathrm{s}$ after outage injection to lap end, the position error settles to $13.4\,\mathrm{cm}$ and $13.2\,\mathrm{cm}$ for the front-thruster cases and $10.9\,\mathrm{cm}$ and $11.4\,\mathrm{cm}$ for the starboard-side cases.

\begin{figure}[!ht]
    \centering
    \includegraphics[width=\linewidth]{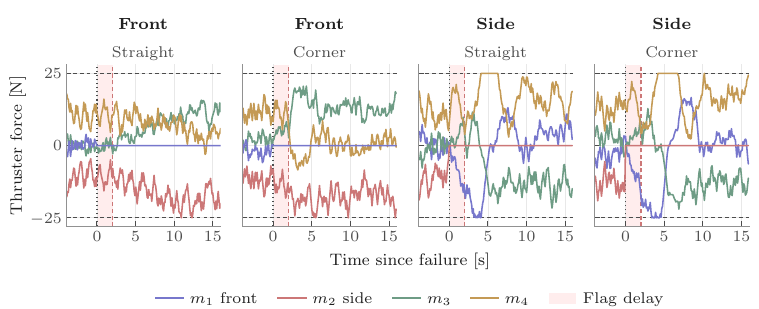}
    \caption{Per-thruster commands for four delayed mid-lap outages.}
    \label{fig:fault_onset_actuation}
    \par\smallskip

\end{figure}

Figure~\ref{fig:fault_onset_actuation} shows the selected thruster command at zero from outage injection. The controller retains its nominal four-thruster model until the flag arrives, when the constraint update removes that thruster from command allocation and redistributes the commanded load among survivors.

\begin{figure}[H]
    \centering
    \includegraphics[width=\linewidth]{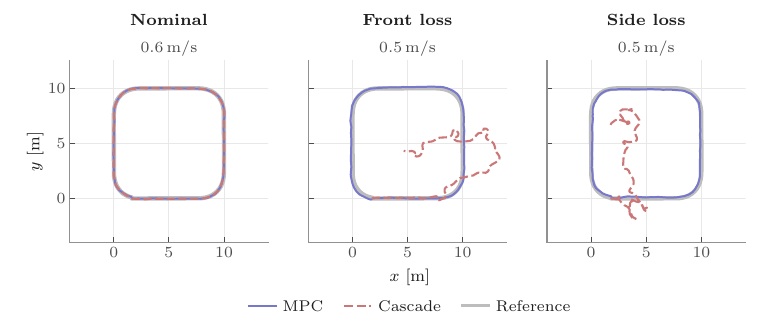}
    \caption{Square trajectories for the \ac{MPC} and cascaded baseline.}
    \label{fig:baseline_xy}
    \par\smallskip
\end{figure}

\begin{figure}[H]
    \centering
    \includegraphics[width=\linewidth]{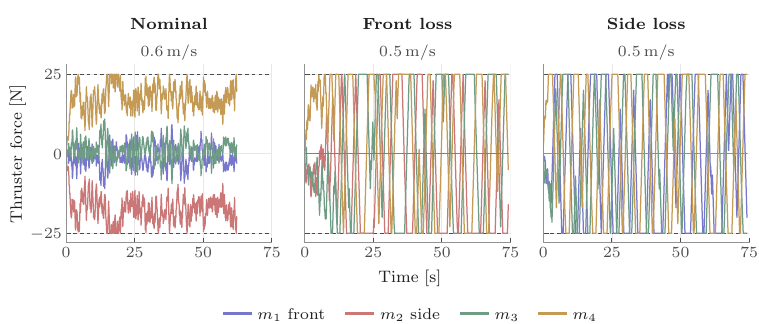}
    \caption{Per-thruster commands for the cascaded baseline.}
    \label{fig:baseline_actuation}
    \par\smallskip
\end{figure}

\subsection{Comparison with the Cascaded Baseline}
\label{sec:baseline_nominal}

The cascade of Sect.~\ref{sec:baseline} was run in the same session, on the same references, and under the same field conditions as far as the experiment permitted. Each comparison below is one paired run. %

Without a failure, the \ac{MPC} achieves the lower reported position-error \ac{RMS} in both station keeping ($1.6$ versus $3.0\,\mathrm{cm}$) and nominal-square tracking ($4.3$ versus $4.7\,\mathrm{cm}$), which are $47\%$ and $9\%$ below the cascaded baseline (Table~\ref{tab:error_metrics}; Fig.~\ref{fig:baseline_xy}). The station-hold error and representative commands in Fig.~\ref{fig:hold_error} provide the corresponding nominal actuator comparison.

For station keeping, the cascade uses $8.7\,\mathrm{N}$ of reported total thruster force versus $5.1\,\mathrm{N}$ for the \ac{MPC}. The largest individual commands are $16.0\,\mathrm{N}$ and $8.8\,\mathrm{N}$, respectively.

Under a single immediate-detection failure, the cascade's reported position-error \ac{RMS} exceeds the \ac{MPC}'s by a factor of $51$ for the front-thruster case and $53$ for the starboard-side case (Table~\ref{tab:error_metrics}). 
Figure~\ref{fig:baseline_xy} shows the resulting contrast: the \ac{MPC} remains near the square reference in both cases, whereas the cascade trajectories depart from it. Figure~\ref{fig:baseline_actuation} shows that the cascade's surviving thrusters repeatedly reach both software command bounds after the failure. The corresponding saturation fractions are $31\%$ and $29\%$ for the cascade, compared with $0.2\%$ and $0.5\%$ for the \ac{MPC} (Table~\ref{tab:error_metrics}). These traces support the interpretation that the downstream saturation prevents the cascaded tracker from accounting for the available actuator set, but they do not establish an asymptotic result or isolate all architectural differences.

\section{Conclusion}\label{sec:conclusion}
This paper described a deployed control system for choreographed autonomous surface vehicles. On each craft, a single condensed \ac{QP} over the per-thruster commands closes
trajectory tracking, thrust allocation, per-thruster force and rate limits, and
post-detection fault reconfiguration at $10\,\mathrm{Hz}$. Fault information enters the same
program through one binary flag per thruster, and the interface is unchanged in form whether four thrusters or two remain available. Fault detection itself remains external to the controller by design.

The field campaign characterizes the resulting behavior across four
experimental sets. %
Station keeping over ten minutes
yields a position-error \ac{RMS} of $1.6\,\mathrm{cm}$, and the
$10\,\mathrm{m}$ rounded square at $0.6\,\mathrm{m/s}$ yields
$4.3\,\mathrm{cm}$. With a single thruster disabled before motion onset, the
position-error \ac{RMS} increases by a factor of approximately $2.7$, to
$11.6\,\mathrm{cm}$ for front-thruster loss and $11.5\,\mathrm{cm}$ for
starboard-side loss at $0.5\,\mathrm{m/s}$, with at most 
$23.3\,\mathrm{cm}$ error. Under the deployed nominal stage cost, the controller preserves
position while admitting increased heading error. Outages injected mid-lap
under load, with the flag withheld for approximately two seconds to
represent detector latency, produce bounded transients in all four runs. With two of four thrusters
unavailable, the craft continues to track the reference shape under released orientation weights, with a \ac{RMS} of $1.25\,\mathrm{m}$.

In a paired comparison against a cascaded implementation, nominal tracking is equivalent to within a few centimeters ($4.3$ versus $4.7\,\mathrm{cm}$ position-error \ac{RMS}), expected for two controllers tuned with comparable effort on the same craft, while the faulted runs separate by a factor of roughly $50$ in position-error \ac{RMS}. %

\bibliographystyle{styles/bibtex/spmpsci_unsrt}
\bibliography{references}

\end{document}